\documentclass{article}
\usepackage{spconf,amsmath,epsfig}
\usepackage{amsmath}
\usepackage{amssymb}

\usepackage{graphicx}
\usepackage{color}
\usepackage{multirow}
\usepackage{array}
\usepackage{booktabs}  
\usepackage{subfig}

\begin{document}\sloppy

\def\x{{\mathbf x}}
\def\L{{\cal L}}

\title{Unsupervised Triplet Hashing for Fast Image Retrieval}
%
\name{Shanshan Huang, Yichao Xiong, Ya Zhang and Jia Wang}
\address{Shanghai Jiao Tong University, Shanghai, China}

\maketitle

\begin{abstract}
Hashing has played a pivotal role in large-scale image retrieval. With the development of Convolutional Neural Network (CNN), hashing learning has shown great promise. 
But existing methods are mostly tuned for classification, which are not optimized for retrieval tasks, especially for instance-level retrieval. 
In this study, we propose a novel hashing method for large-scale image retrieval. 
Considering the difficulty in obtaining labeled datasets for image retrieval task in large scale, we propose a novel CNN-based unsupervised hashing method, namely Unsupervised Triplet Hashing (UTH). 
The unsupervised hashing network is designed under the following three principles: 
1) more discriminative representations for image retrieval; 
2) minimum quantization loss between the original real-valued feature descriptors and the learned hash codes; 
3) maximum information entropy for the learned hash codes.  
Extensive experiments on {\bfseries{CIFAR-10}}, {\bfseries{MNIST}} and {\bfseries{In-shop}} datasets have shown that UTH outperforms several state-of-the-art unsupervised hashing methods in terms of retrieval accuracy. 
\end{abstract}
\section{Introduction}
\label{sec:intro}
With the explosive growth of multimedia contents, how to speed up image retrieval draws much attention in computer vision.
Hashing, which uses mapping functions to transform a high-dimensional feature vector into a compact and expressive binary codes \cite{har2012approximate, gong2011iterative, weiss2009spectral}, has shown significant success for fast image retrieval. 
In recent years, with the rapid development of Convolutional Neural Network (CNN), several CNN-based hashing  methods~\cite{lin2015, xia2014supervised, lin2015deephash, wang2012semi,salakhutdinov2009semantic, lin2015tiny,linlearning2016} have been proposed and demonstrated promising results. 
In particular, unsupervised hashing learning has recently received increasing attention because it does not require labeled training data thus making the methods widely applicable. 
The earliest studies use stacked Restricted Boltzmann Machines (RBMs) to encode binary codes~\cite{salakhutdinov2009semantic, lin2015tiny} for unsupervised hashing. 
However, RBMs are complex and require pre-training, which are not efficient for practical applications. 
More recently, data augmentation is leveraged to reinforce the representation ability of the model \cite{linlearning2016}, which achieves the state-of-the-art results so far. 
They augment the training data with different rotations of the reference images, and attempt to minimize the distances between the binary codes of the reference images and that of their rotations. 
However, optimizing with rotation invariance between images and their rotations only provides positive data for the learning, and cannot guarantee the model to generate discriminative binary codes for different images. 
\begin{figure}
\begin{center}
  \includegraphics[height=3.2cm]{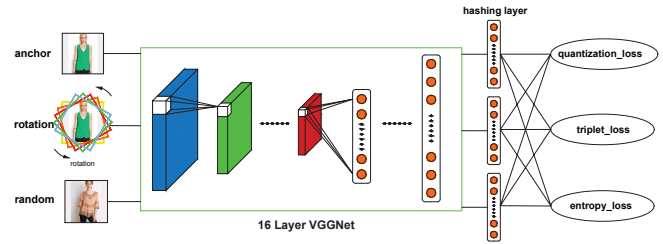}
  \caption{The proposed UTH method. We add a hashing layer to generate a compact feature vector. The input to our architecture is an image triplet consisting of an anchor image, a rotated image and a random image. These image triplets share parameters in our architecture. At the top of our architecture, we use three criterions to learn efficient codes: 
1) more discriminative representations for image retrieval; 
2) minimum quantization loss between the original real-valued feature descriptors and the learned hash codes; 
3) maximum information entropy for the learned hash codes. }
  \label{Fig:model}
\end{center}
\end{figure}

In this paper, we propose a novel unsupervised hashing method, namely Unsupervised Triplet Hashing (UTH), based on CNN with triplet loss to ensure the discriminability of the hash codes. 
The novelties of this paper are 1) we replace the rotation invariance loss in DeepBit \cite{linlearning2016}, one of the state-of-the arts deep hashing method, with a triplet loss; 
2) We construct triplets from unlabeled data in optimizing the triplet loss. 
The triplet loss enforces balanced training samples while the rotation invariance loss fails to consider negative training samples.
The three key components of UTH are illustrated in Fig.~\ref{Fig:model}. 
First of all, we design a triplet loss to learn more discriminative representations for fast image retrieval, which enforces a margin between the distances of rotated images and random images to the original images. 
This allows the rotated images for each image to live on a manifold, while still enforcing the distances and thus discriminability to other images. 
Then we add two constraints to guarantee the retrieval performance of the learned hash codes, a minimum quantization loss between the original real-valued feature descriptors and the learned hash codes to maintain the high retrieval accuracy and a maximum information entropy loss to reinforce the representation ability of learned hash codes. 
We evaluate the proposed UTH architecture on three benchmarks, and show that it outperforms several state-of-the-art unsupervised hashing methods in terms of retrieval accuracy. 

\section{Related Works}\label{sec:Related Works}
\textbf{Hashing Method.} According to whether the semantic information is used, learning-based hashing method can be divided into three categories: supervised hashing~\cite{lin2015,xia2014supervised,lin2015deephash,nguyen2016deep,lai2015simultaneous}, semi-supervised hashing~\cite{wang2012semi} and unsupervised hashing~\cite{har2012approximate,gong2011iterative,weiss2009spectral,salakhutdinov2009semantic,lin2015tiny,linlearning2016}. 

For supervised hashing, 
Lin \emph{et al.}~\cite{lin2015} employ a hidden layer to learn binary hash codes by using a ``Softmax'' layer on the top of the model. The learned features which are optimized for image classification are directly applied to image retrieval. 
Two-step learning method is adopted to learn binary hash codes. 
For example, Xia \emph{et al.}~\cite{xia2014supervised} adopt a learning method which learns binary codes for all the training data in the first step and learns hash functions on the basis of the learned codes in the second step. 
Lin \emph{et al.}~\cite{lin2015deephash} introduce a hashing scheme based on stacked RBMs and Siamese network, in which stacked RBMs are aimed to learn the initial parameters of the network and then the parameters are fine-tuned through a Siamese network. 
Nguyen \emph{et al.}~\cite{nguyen2016deep} use a triplet loss function to minimize the Hamming distance between the neighbor pairs while preserving the relative similarity of non-neighbor pairs with a relaxed empirical penalty. 
Lai \emph{et al.}~\cite{lai2015simultaneous} present a divide-and-encode module to divide the intermediate image features into multiple branches, each encoded into one hash bit, then use a triplet loss to fine-tune the network. 

For semi-supervised hashing, Wang \emph{et al.}~\cite{wang2012semi} present a semi-supervised hashing (SSH) framework to learn hash codes by minimizing empirical error on the labeled data and maximizing variance and independence of hash codes over the labeled and unlabeled data. 

For unsupervised hashing, most of the previous unsupervised methods~\cite{har2012approximate,gong2011iterative,weiss2009spectral} make use of hand-crafted image features and are not end-to-end. 
Har-Peled \emph{et al.}~\cite{har2012approximate} propose Local Sensitive Hashing (LSH), which uses random projections to construct hash functions, making samples within short Hamming distance in hash space be near in their source space. 
Gong \emph{et al.}~\cite{gong2011iterative} propose the popular Iterative Quantization (ITQ), which first performs PCA and then learns a rotation to minimize the quantization error of mapping the transformed data to the vertices of a zero-centered binary hypercube. 
As the deep learning develops, many unsupervised hashing methods~\cite{salakhutdinov2009semantic,lin2015tiny,linlearning2016} based on deep learning are proposed. 
Salakhutdinov, \emph{et al.}~\cite{salakhutdinov2009semantic} propose semantic hashing (SH), which uses RBMs as an auto-encoder network to generate efficient binary codes. 
Lin \emph{et al.}~\cite{linlearning2016} propose  DeepBit to learn a set of nonlinear mapping functions by inserting a latent layer into the previous model and construct pair-wise training data by combining the original images with its rotated images, which outperforms state-of-the-art unsupervised schemes. 

\textbf{Deep Learning.} Recently, deep learning has achieved explosive success in pattern recognition including image classification, segmentation and learning-based hashing for fast image retrieval. 
Guo \emph{et al.}~\cite{guo2015cnn} propose a straightforward CNN-based hashing method, they quantize the activations of a fully connected layer with threshold $0$ and take the binary result as hash codes. 
Liong \emph{et al.}~\cite{erin2015} present a framework to learn binary codes by seeking multiple hierarchical non-linear transformations, so that the nonlinear relationship of samples can be well exploited. 
Xia \emph{et al.}~\cite{xia2014supervised} present a framework to automatically learn a good image representation tailored to hashing as well as a set of hash functions. 
Yao \emph{et al.}~\cite{yaodeep} propose a co-training hashing network by jointly learning projections from image representations to hash codes and classification. 

\section{The proposed approach}

Generally, the proposed UTH architecture contains three major components:
1) learning more discriminative representations for image retrieval via a triplet loss;
2) minimizing quantization loss between the original real-valued feature descriptors and the learned hash codes to maintain the high retrieval performances;
3) maximizing information entropy for the learned hash codes to carry as much information as possible.
The whole architecture is shown in Fig.~\ref{Fig:model}. Let $\mathcal{L}_{T}$ denote the triplet loss function, $\mathcal{L}_{Q}$ denote the quantization loss function and $\mathcal{L}_{E}$ denote the entropy loss function.
We define an overall loss function:
\begin{align}\label{equ:total_loss_function}
\mathcal{L} = \alpha\mathcal{L}_{T} + \beta\mathcal{L}_{Q} + \gamma \mathcal{L}_{E},
\end{align}
where $\alpha,\beta$ and $\gamma$ are the parameters for each object. 
These loss functions will be explained in the following chapters.

\subsection{Unsupervised Triplet Loss}

To ensure the discriminability of the hash codes, we propose an unsupervised triplet neural network. 
A triplet training set is constructed by from the unlabeled data in the following means. 
For each image in the unlabeled set, a rotation of the image, a randomly selected image from the dataset, and itself form a triplet. 
It is safe to assume that the distance between the rotation of the image to the image is smaller than that of the randomly selected image to the image. 

Let $(p, p^+, p^-)$ denote a triplet example. 
$\mathcal{F}()$ denotes the hashing function we have learned. 
Specifically, $\mathcal{F}(p)$ is the feature of the anchor image, $\mathcal{F}(p^+)$ and $\mathcal{F}(p^-)$ are the features of the rotated image and the random image respectively. 
The triplet loss function is written as
\begin{align}\label{eqn:triplet}
&\mathcal{L}_{T} \nonumber\\
=& \max\{0, m + \mathcal{D_E}(\mathcal{F}(p), \mathcal{F}(p^+)) - \mathcal{D_E}(\mathcal{F}(p), \mathcal{F}(p^-))\} \nonumber \\
=& \max\{0, m + \|\mathcal{F}(p)-\mathcal{F}(p^+)\|{_2^2} - \|\mathcal{F}(p)-\mathcal{F}(p^-)\|{_2^2}\},
\end{align}
where $\mathcal{D_E}$ denotes the Euclidean distance between two objects and we use $L_2$-norm to calculate the distance, and $m$ denotes the margin we select in our method.

In preparing training dataset, we rotate each image $p$ in the training set by some fixed degree to form a $p^+$, and randomly select an image except itself to form a $p^-$, which constructs a triplet $(p, p^+, p^-)$.

\subsection{Quantization Loss}
In order to learn multiple nonlinear hashing functions, we add an activation layer followed by the hashing layer.   
In our study, ReLU is chosen as the activation function because it prevents from gradient disappearance in our training process. 
A binary hash code is generated by quantizing the output feature. 
The quantization rule is shown as
\begin{equation}\label{eqn:min quantization loss}
b = \begin{cases}
1, &{\text{if}\ \mathcal{F}(p) > threshold}, \\
0, &\text{otherwise}.
\end{cases}
\end{equation}

In our experiments, we set the threshold as $0.5$ and add a constraint to narrow the gap between the retrieval performances before and after quantizing the image features.
The minimum quantization loss, i.e., $\mathcal{L}_{Q}$, is defined as
\begin{align}\label{eqn:min quantization loss}
\mathcal{L}_{Q} =& \sum_{n=1}^N\sum_{m=1}^M\|\mathcal{F}(p)-b\|^2 \nonumber \\
=& \begin{cases}
\sum\limits_{n=1}^N\sum\limits_{m=1}^M\|\mathcal{F}(p)-1\|^2, &{\text{if}\ \mathcal{F}(p) > 0.5}, \\
\sum\limits_{n=1}^N\sum\limits_{m=1}^M\|\mathcal{F}(p)\|^2, &\text{otherwise},
\end{cases}
\end{align}
where $N$ is the number of training data, $M$ is the length of hash codes.

The loss function \eqref{eqn:min quantization loss} pushes the real value of each dimension to either 0 or 1, thus the retrieval performance by using quantized image features, i.e. hash codes, is approximate to the performance by using real-value image features.
\subsection{Entropy Loss}
According to information theory, the highest entropy is reached when information distributes evenly among each bit in the code.
Thus a higher entropy means that the code carries more information.
Inspired by this theory and DeepBit, we add a constraint to impel each bit in our output binary codes to be evenly distributed. Hence, the maximum entropy loss is formulated as
\begin{equation}\label{eqn:max entropy}
\mathcal{L}_{E} = \sum_{m=1}^{M}(\mu_m-0.5)^2,\\
\mu_m = \frac{1}{N}\sum_{n=1}^Nb_n(m).
\end{equation}
%
%

Substituting equations~\eqref{eqn:triplet},~\eqref{eqn:min quantization loss}, and~\eqref{eqn:max entropy} into equation~\eqref{equ:total_loss_function}, we can obtain the overall loss function.

\section{Experiments}
\subsection{Experimental Setting}
To test the generalizability of different hashing methods, we choose three public datasets of different characteristics to evaluation the methods under comparison, i.e., CIFAR-10~\cite{krizhevsky2009learning}, MNIST\footnote{http://yann.lecun.com/exdb/mnist/.} and In-shop~\cite{liu2016deepfashion}. 
The basic statistic of the datasets is shown in Table~\ref{tab:dataset}. 
The CIFAR-10 dataset is chosen for a direct comparison with DeepBit. 
MNIST, a dataset quite different from ImageNet which is used for the pre-trained model, is selected to test the generalizability of different hashing methods. 
Finally, the In-shop dataset is chosen to test different hashing methods for instance-level retrieval.

%
%
\begin{table*}[t]
\begin{center}
\caption{The basic statistic information of the selected datasets in our experiments.} \label{tab:dataset}
\vspace{0.08in}
\begin{tabular}{| c | c | c | c | c |}
\hline
 Dataset & Image Number & Category & Training Number & Testing Number \\
 \hline
 CIFAR-10~\cite{krizhevsky2009learning} & 60,000 & 10 & 50,000 & 10,000 \\
 \hline
 MNIST \footnotemark[1] & 70,000 & 10 & 60,000 & 10,000 \\
 \hline
 In-shop~\cite{liu2016deepfashion} & 52,712 & 7,982 & 38,494 & 14,218 \\
\hline
\end{tabular}
\end{center}
\end{table*}
%
We compare the proposed UTH method with the state-of-the-art unsupervised hashing methods: KMH~\cite{he2013k}, SphH~\cite{heo2012spherical}, SpeH~\cite{weiss2009spectral}, PCAH~\cite{wang2012semi}, LSH~\cite{har2012approximate}, PCA-ITQ~\cite{gong2011iterative}, DH~\cite{erin2015}, DeepBit~\cite{linlearning2016}. 
Similar to previous studies,  we evaluate the performance of the hashing methods with the following two widely-adopted metrics: mean average precision (mAP) at top 1,000 and Recall-Precision curve. 
%
\begin{figure*}[htb]
\centering
\subfloat[16 bits]{
\begin{minipage}[t]{0.32\linewidth}
\centering
\includegraphics[width=5.6cm]{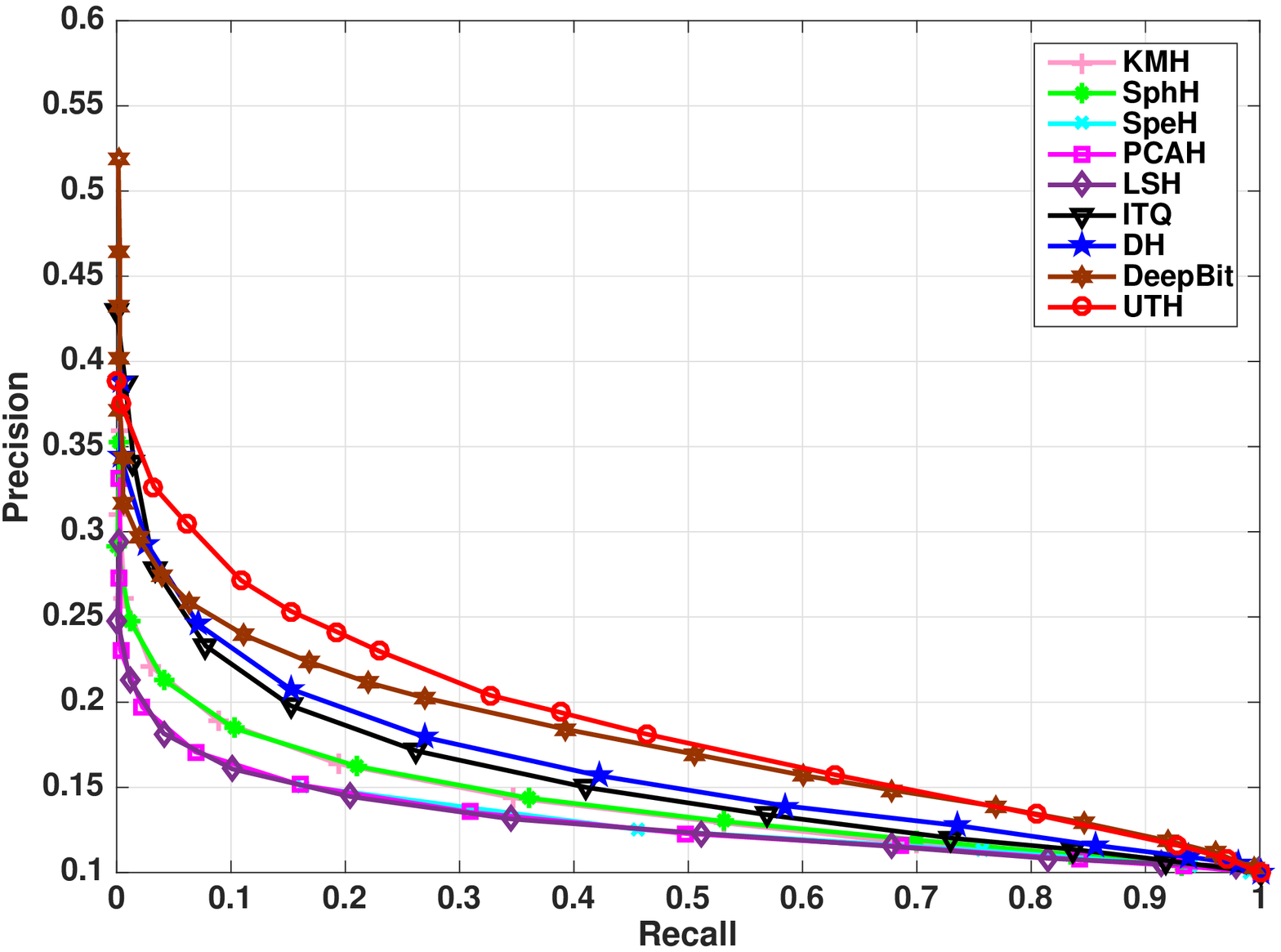}\\
\end{minipage}
}
\subfloat[32 bits]{
\begin{minipage}[t]{0.32\linewidth}
\centering
\includegraphics[width=5.6cm]{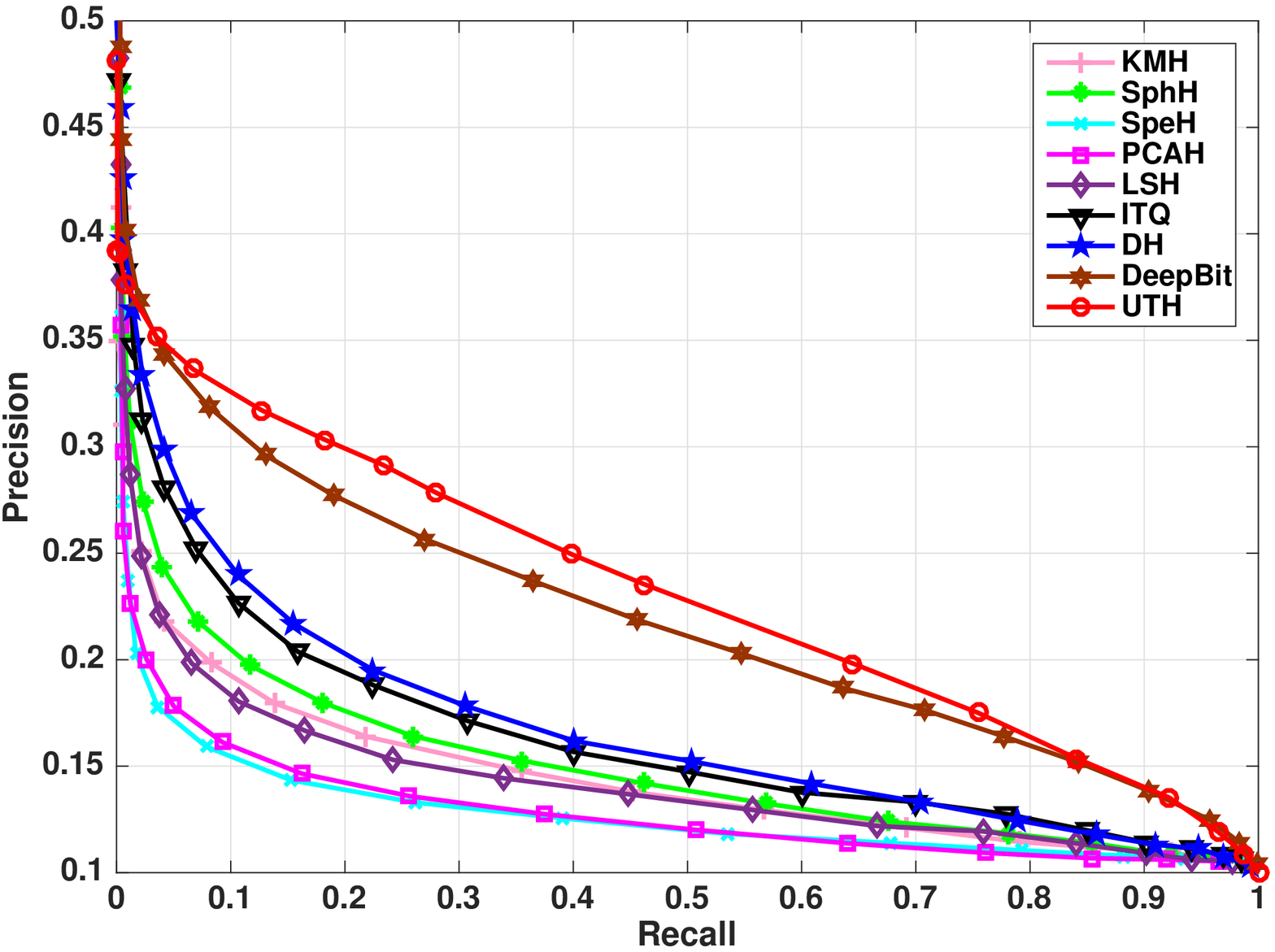}\\
\end{minipage}
}
\subfloat[64 bits]{
\begin{minipage}[t]{0.32\linewidth}
\centering
\includegraphics[width=5.6cm]{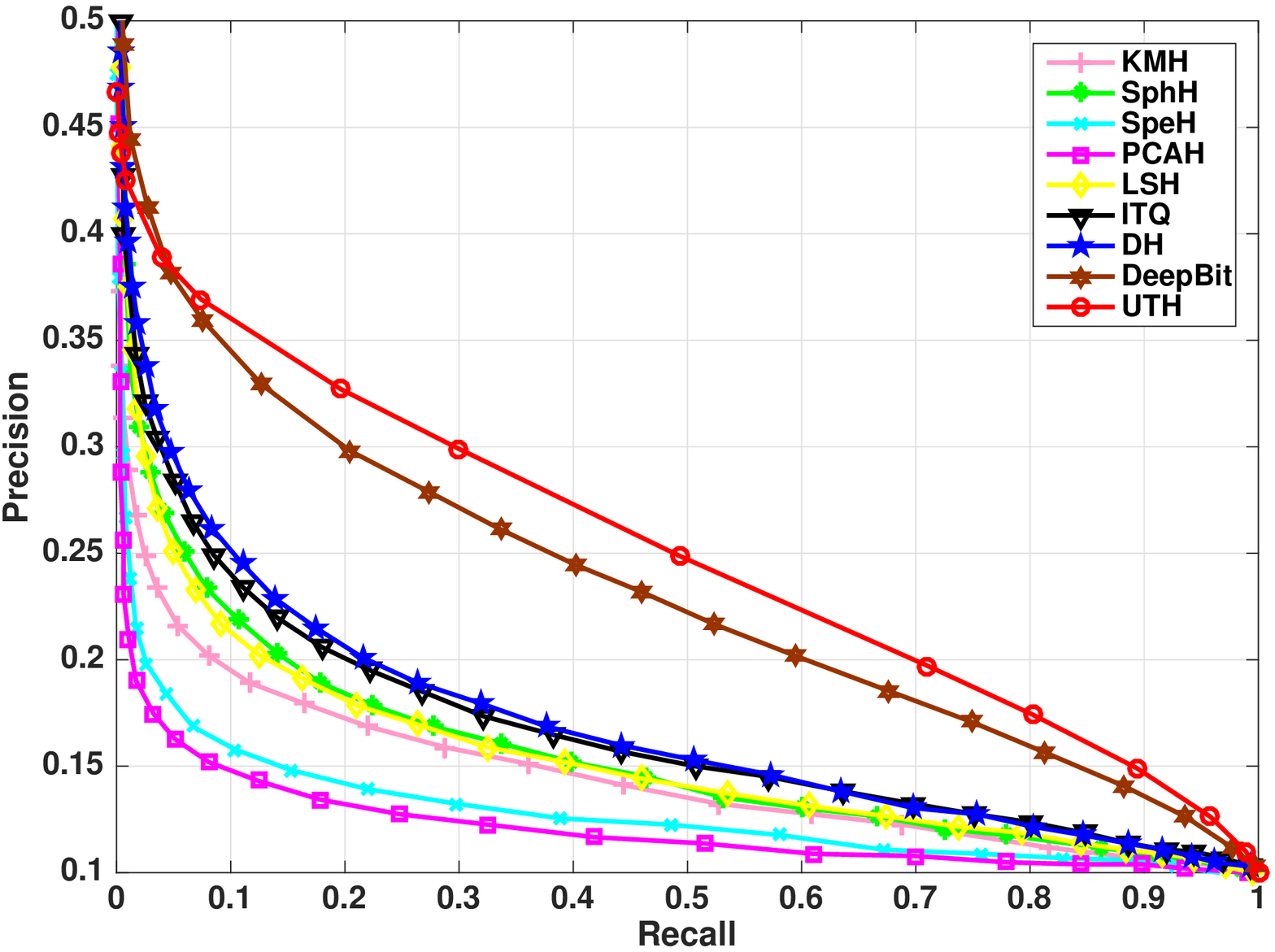}\\
\end{minipage}
}
\caption{Recall-Precision curves on the CIFAR-10 dataset for different unsupervised hashing methods with respect to 16, 32 and 64 bits, respectively.}
\label{Fig:precision_recall_cifar10}
\end{figure*}

We implement UTH using the open source Caffe~\cite{jia2014caffe} and update parameters by Stochastic Gradient Descent. 
Besides, we use VGGNet~\cite{simonyan2014very} in UTH for pre-training and add a latent layer (named hashing layer) following by a ReLU activation layer. 
We adopt the idea of using weight sharing network for model fine-tuning to learn a more generalized network. 
When generating $p^+$, we rotate each image by $-10,-5,5,10$ degrees respectively. 
In training process, we set $\alpha=\beta=\gamma=1$, and follow the same setting as DeepBit. 
We update the parameters of the network by minimizing the quantization error and entropy error using the original training data at first, then we fine-tune the network by adding a triplet loss using the triplet examples we have constructed. 
We set the output hashing layer as 16 bits, 32 bits and 64 bits respectively for CIFAR-10 and MNIST datasets. 
Considering the complexity of In-shop dataset, we set the output hashing layer as 64 bits, 128 bits and 256 bits. 
In comparison with the other unsupervised hashing methods, the results of the compared methods are from~\cite{linlearning2016,erin2015} for CIFAR-10 and MNIST datasets, 
and the results of the traditional methods are obtained by representing each image as a 512-D GIST feature for In-shop dataset. 
\subsection{Experiments Results}
\begin{table}[t]
\begin{center}
\caption{Mean Average Precision (mAP, \%) at top 1,000 of different unsupervised hashing methods on CIFAR-10 dataset with respect to different hash codes.} \label{tab:map_cifar10}
\vspace{0.08in}
\begin{tabular}{p{2cm}p{1.6cm}p{1.6cm}p{1cm}}
\toprule[1pt]
 Method & 16-bit & 32-bit & 64-bit \\
 \hline
  KMH~\cite{he2013k} & 13.59 &13.93 & 14.46 \\
  SphH~\cite{heo2012spherical} & 13.98 & 14.58 &15.38 \\
  SpeH~\cite{weiss2009spectral} & 12.55 & 12.42 & 12.56  \\
  PCAH~\cite{wang2012semi} & 12.91 & 12.60 & 12.10 \\
  LSH~\cite{har2012approximate} & 12.55 & 13.76 & 15.07 \\
  PCA-ITQ~\cite{gong2011iterative} & 15.67 & 16.20 & 16.64 \\
  DH~\cite{erin2015} & 16.17 & 16.62 & 16.69 \\
  DeepBit~\cite{linlearning2016} & 19.43 & 24.86 & 27.73 \\
  \hline
  {\bfseries{UTH}} & {\bfseries{28.66}} & {\bfseries{30.66}} & {\bfseries{32.41}} \\
  \bottomrule[1pt]
\end{tabular}
\end{center}
\end{table}
\begin{table}[t]
\begin{center}
\caption{Mean Average Precision (mAP, \%) at top 1,000 of different unsupervised hashing methods on MNIST dataset with respect to different hash codes.} \label{tab:map_mnist}
\vspace{0.08in}
\begin{tabular}{p{2cm}p{1.6cm}p{1.6cm}p{1cm}}
\toprule[1pt]
 Method & 16-bit & 32-bit & 64-bit \\
  \hline
  KMH~\cite{he2013k} & 32.12 & 33.29 & 35.78 \\
  SphH~\cite{heo2012spherical} & 25.81 & 30.77 & 34.75 \\
  SpeH~\cite{weiss2009spectral} & 26.64 & 25.72 & 24.10 \\
  PCAH~\cite{wang2012semi} & 27.33 & 24.85 & 21.47 \\
  LSH~\cite{har2012approximate} & 20.88 & 25.83 & 31.71 \\
  PCA-ITQ~\cite{gong2011iterative} & 41.18 & 43.82 & 45.37 \\
  DH~\cite{erin2015} & 43.14 & 44.97 & 46.74 \\
  DeepBit~\cite{linlearning2016} & 28.18 & 32.02 & 44.53 \\
  \hline
  {\bfseries{UTH}} & {\bfseries{43.15}} & {\bfseries{46.58}}& {\bfseries{49.88}} \\
 \bottomrule[1pt]
\end{tabular}
\end{center}
\end{table}
\begin{figure*}[htb]
\centering
\subfloat[16 bits]{
\begin{minipage}[t]{0.32\linewidth}
\centering
\includegraphics[width=5.6cm]{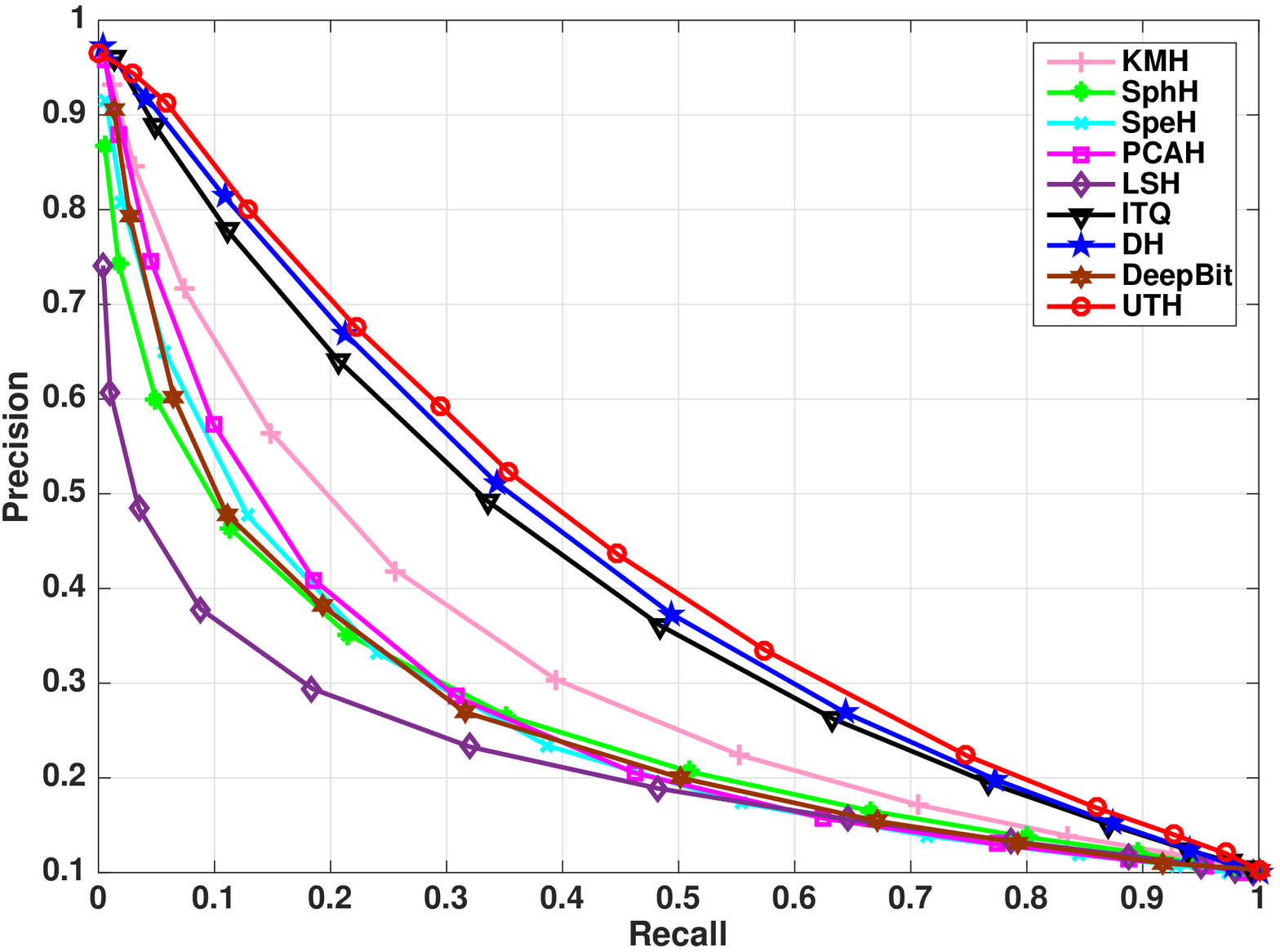}\\
\end{minipage}
}
\subfloat[32 bits]{
\begin{minipage}[t]{0.32\linewidth}
\centering
\includegraphics[width=5.6cm]{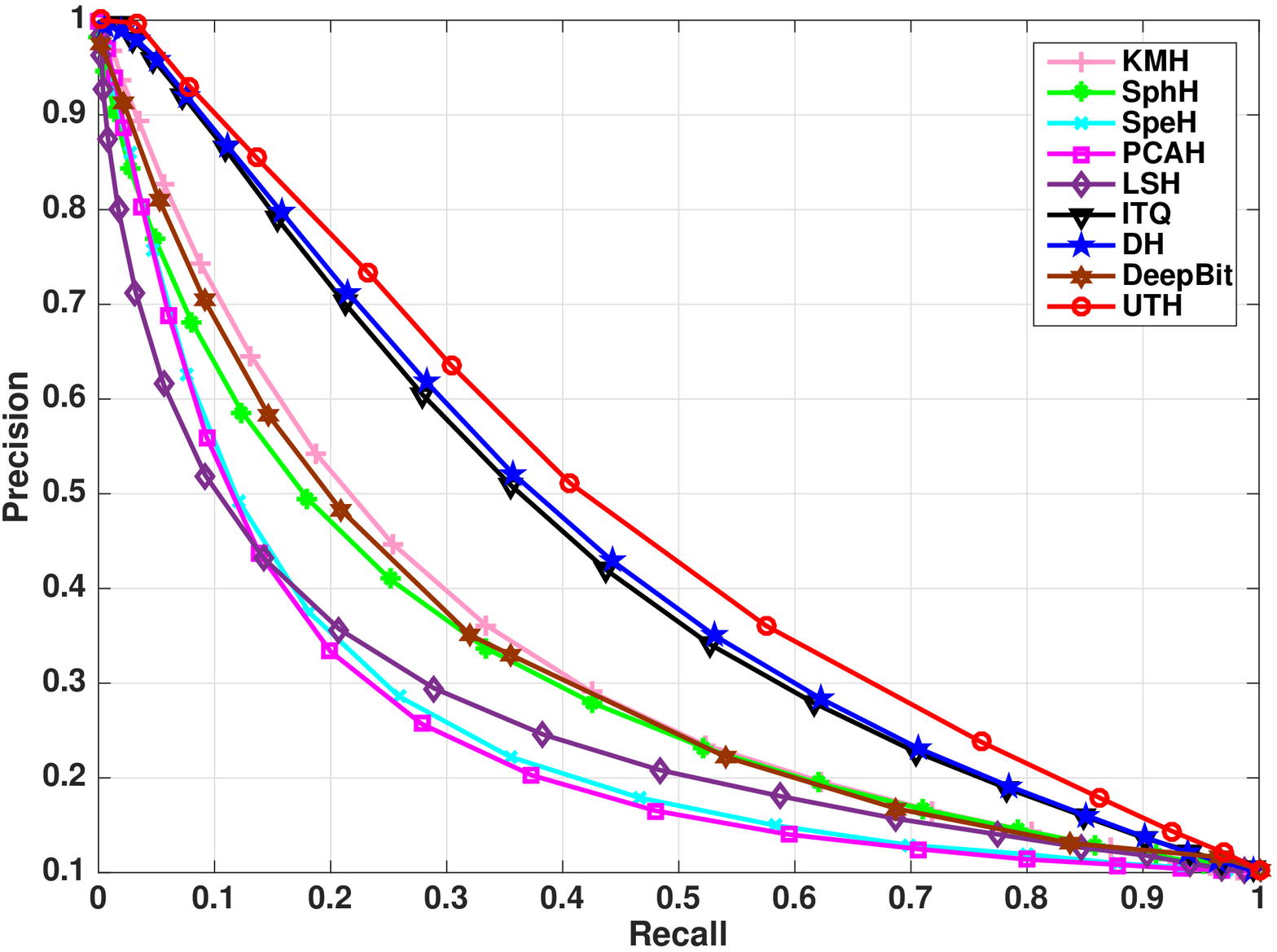}\\
\end{minipage}
}
\subfloat[64 bits]{
\begin{minipage}[t]{0.32\linewidth}
\centering
\includegraphics[width=5.6cm]{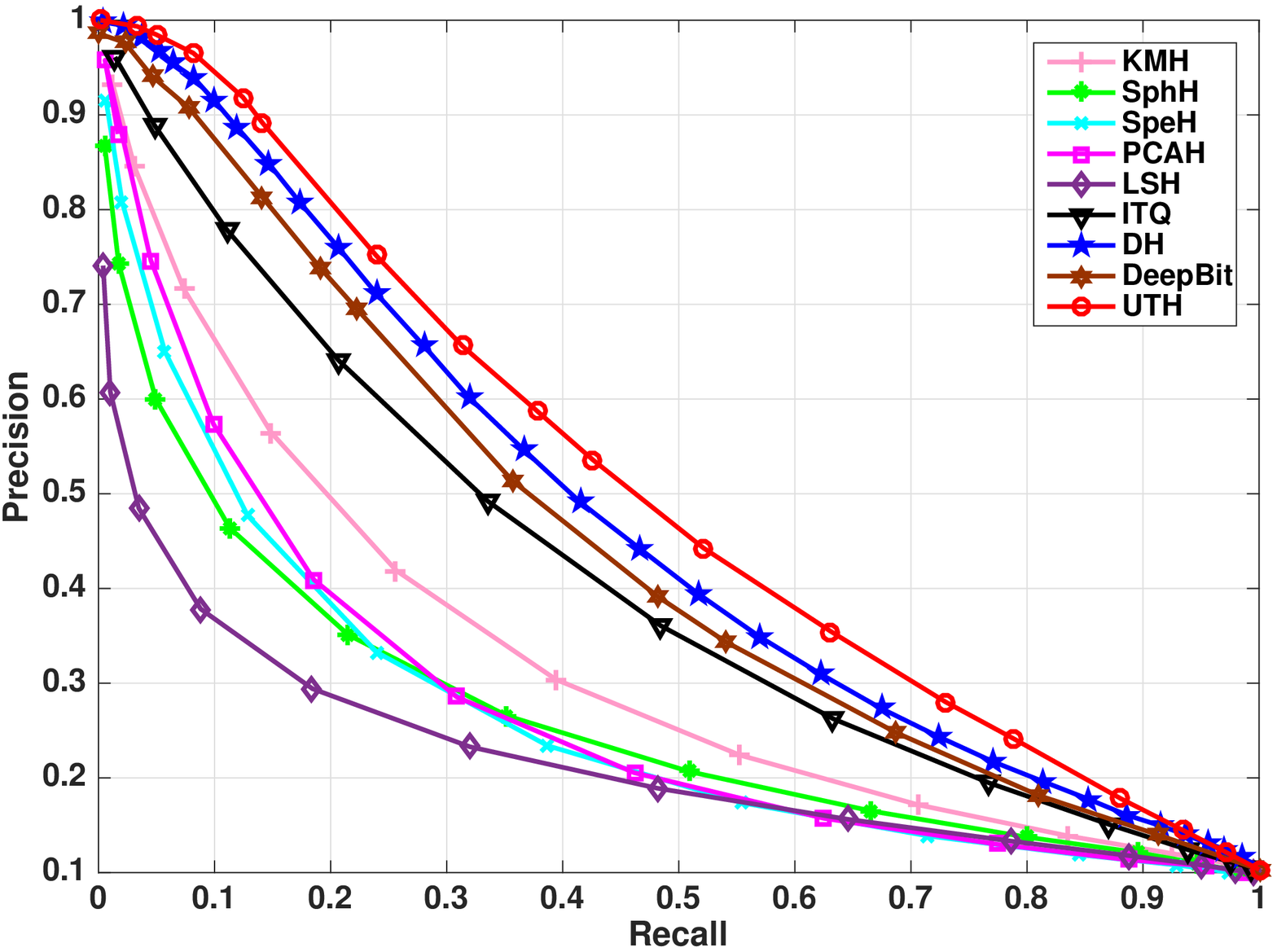}\\
\end{minipage}
}
\caption{Recall-Precision curves on the MNIST dataset for different unsupervised hashing methods with respect to 16, 32 and 64 bits, respectively.}
\label{Fig:precision_recall_mnist}
\end{figure*}
\begin{figure*}[htb]
\centering
\subfloat[64 bits]{
\begin{minipage}[t]{0.32\linewidth}
\centering
\includegraphics[width=5.6cm]{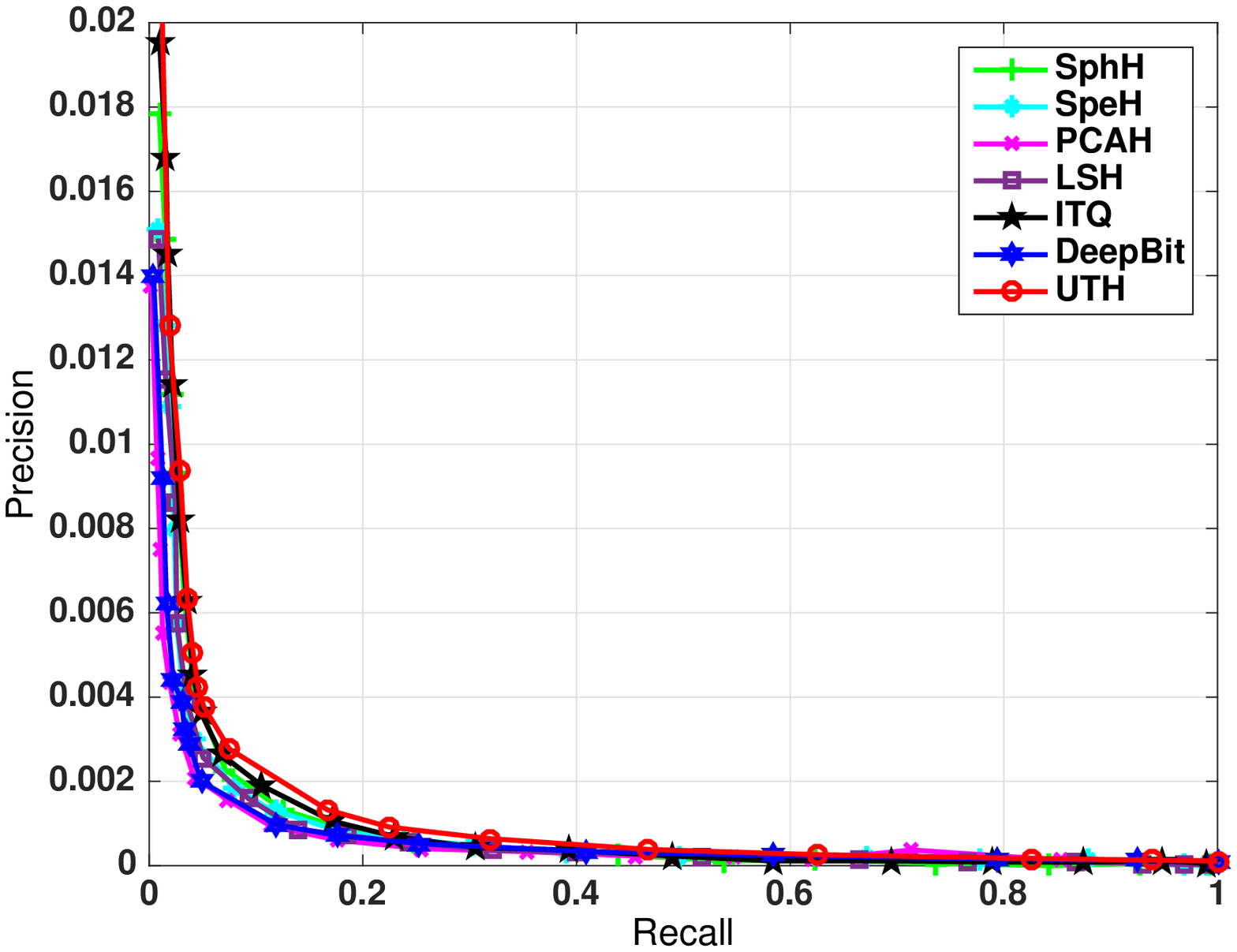}\\
\end{minipage}
}
\subfloat[128 bits]{
\begin{minipage}[t]{0.32\linewidth}
\centering
\includegraphics[width=5.6cm]{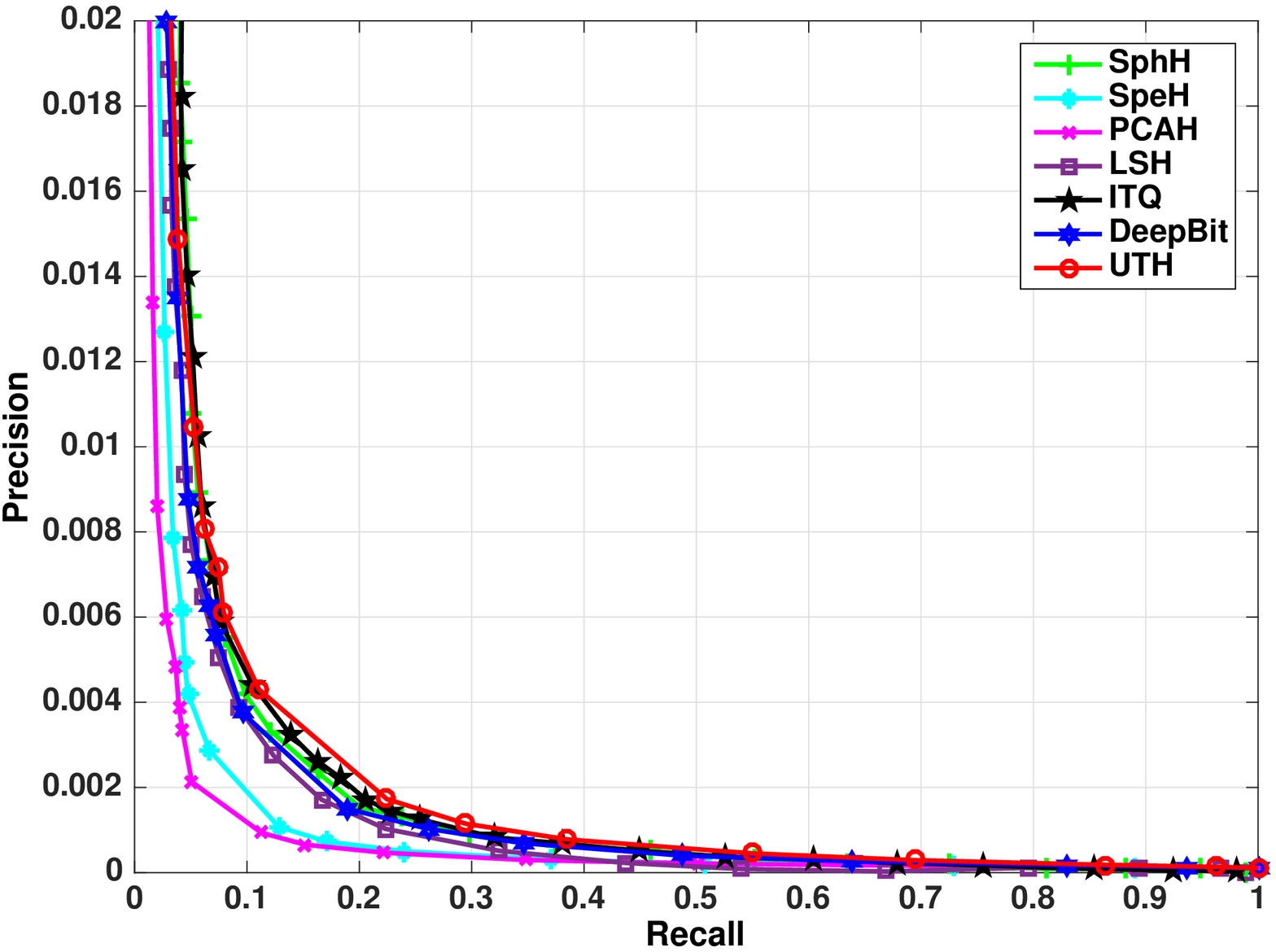}\\
\end{minipage}
}
\subfloat[256 bits]{
\begin{minipage}[t]{0.32\linewidth}
\centering
\includegraphics[width=5.6cm]{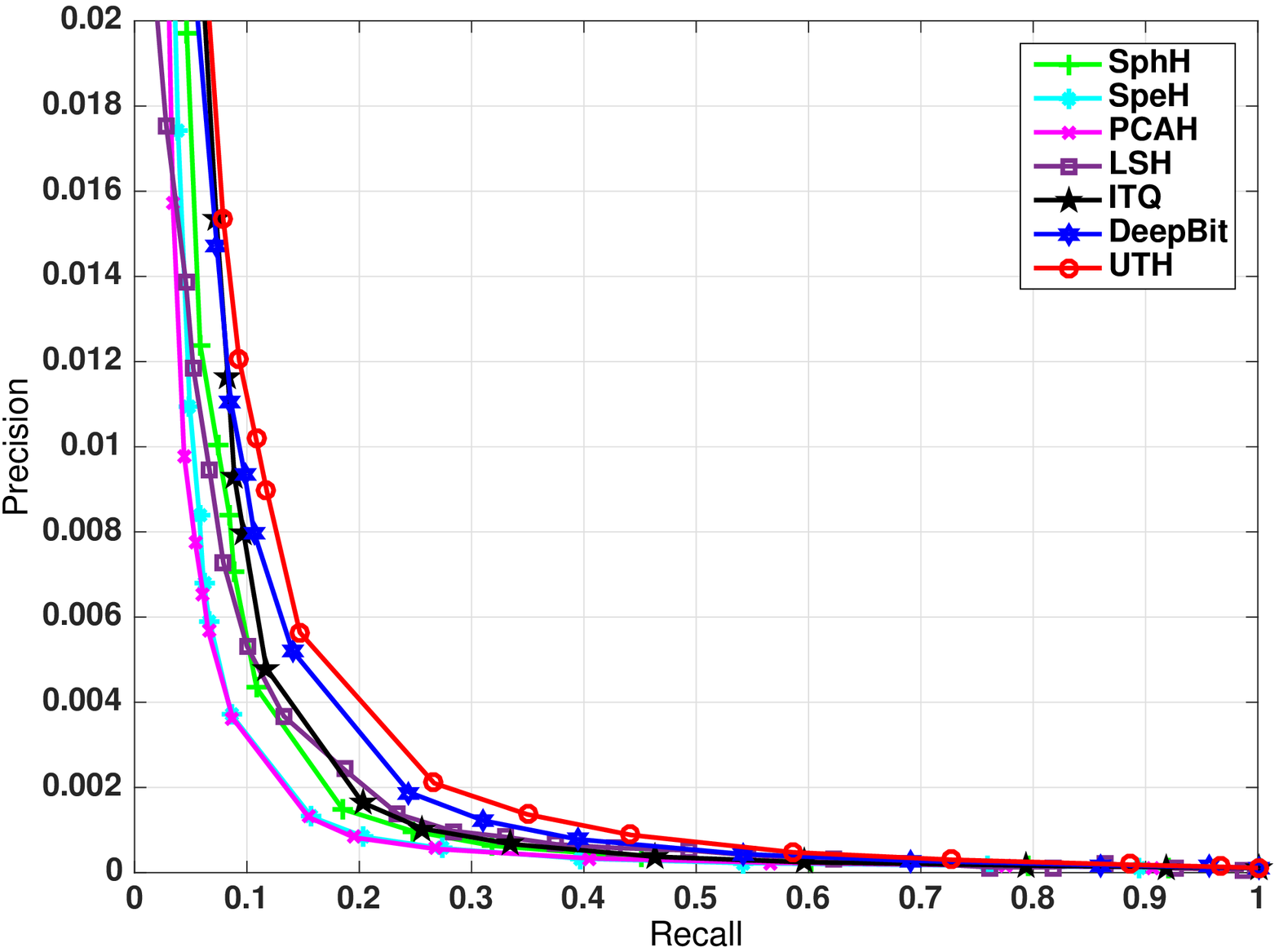}\\
\end{minipage}
}
\caption{Recall-Precision curves on the In-shop dataset for different unsupervised hashing methods with respect to 64, 128 and 256 bits, respectively.}
\label{Fig:precision_recall_inshop}
\end{figure*}
Following DeepBit, we randomly sample 1000 images as the query data, and use the remaining images as the gallery set for each dataset. 
Table~\ref{tab:map_cifar10}, Table~\ref{tab:map_mnist} and Table~\ref{tab:map_In-shop} show the mAP results at top 1,000 of different unsupervised hashing methods. 
The Recall-Precision curves are presented Fig.~\ref{Fig:precision_recall_cifar10}, Fig.~\ref{Fig:precision_recall_mnist} and Fig.~\ref{Fig:precision_recall_inshop}. 
For all the three datasets, the proposed UTH outperforms the other unsupervised hashing methods under comparison. 
The consistently better performance of UTH has demonstrated that the proposed UTH learns more discriminative hash codes for fast image retrieval. 

From the results of mAP, UTH improves the retrieval accuracy with respect to 16-bit, 32- bit and 64-bit hash codes by $9.23\%$, $5.80\%$, $4.68\%$ on CIFAR-10 dataset, $14.97\%$, $14.56\%$, $5.33\%$ on MNIST dataset and $3.87\%$, $2.40\%$, $0.61\%$ on In-shop respectively by comparing with DeepBit \cite{linlearning2016}. 
That means the proposed UTH achieves even better performance on more compact hash codes compared with DeepBit. 

The significant improvement of UTH over DeepBit lies partly in that DeepBit optimizes the rotation invariance between images and their rotations, which is to provide rotation invariant descriptors for images. 
As a result, it may not guarantee the model to generate discriminative binary codes for different images. 
With UTH, we attempt to maximize the discriminability of the hash codes while still keeping the rotation invariant features using a triplet network. 
UTH learns weights of all layers in order to consider the co-adaption between neighboring layers in CNNs which has been proved important in \cite{yosinski2014transferable}, while DeepBit freezes the parameters of layers lower than the hashing layer when training, which is another reason why UTH is superior to DeepBit. 

The improvement of UTH over DH [24] lies in that our method utilizes the 16 layers VGGNet as the initialized network and fine tune the network by three loss components. DH takes only three layer hierarchical neural networks to learn hash codes. The improvement over the other hashing methods~\cite{he2013k,heo2012spherical,weiss2009spectral,wang2012semi,har2012approximate,gong2011iterative} lies in that our proposed architecture is based on a deep CNN, which is an end-to-end network, which has been proved to have advantage over hashing methods using hand crafted image features.
\begin{table}[t]
\begin{center}
\caption{Mean Average Precision (mAP, \%) at top 20 of different unsupervised hashing methods on In-shop dataset with respect to different hash codes.} \label{tab:map_In-shop}
\vspace{0.08in}
\begin{tabular}{p{2cm}p{1.6cm}p{1.6cm}p{1.2cm}}
\toprule[1pt]
  Method & 64-bit & 128-bit & 256-bit\\
  \hline
  SphH~\cite{heo2012spherical} & 9.03 & 15.52 & 17.94 \\
  SpeH~\cite{weiss2009spectral} & 8.77 & 12.38 & 17.14 \\
  PCAH~\cite{wang2012semi} & 6.60 & 10.32 & 14.73 \\
  LSH~\cite{har2012approximate} & 8.34 & 13.51 & 15.39 \\
  PCA-ITQ~\cite{gong2011iterative} & 9.77 & 16.74 & 21.29 \\
  DeepBit~\cite{linlearning2016} & 6.53 & 14.70 & 22.65 \\
  \hline
  {\bfseries{UTH}} & {\bfseries{10.40}} & {\bfseries{17.10}} & {\bfseries{23.26}} \\
 \bottomrule[1pt]
  \end{tabular}
 \end{center}
\end{table}
\section{Conclusion}

In this paper, we present a novel unsupervised hashing method based on convolutional neural network (CNN) called unsupervised triplet hashing (UTH). 
The UTH is designed with a triplet network structure to simultaneously achieve the following three objectives: 
1) discriminative representations for fast image retrieval; 
2) accurate binary feature descriptors; 
3) maximizing the information of the learned hash codes.
Extensive experiment evaluations based on {\bfseries{CIFAR-10}}, {\bfseries{MNIST}} and {\bfseries{In-shop}} datasets have showed the promise of the proposed UTH method for fast image retrieval.
\bibliographystyle{IEEEbib}
\bibliography{huangss_bib}

\end{document}